\documentclass{article}
\usepackage{spconf,amsmath,graphicx,amssymb}
\usepackage{multirow,diagbox}
\usepackage{epstopdf}
\usepackage[dvipsnames]{xcolor}
\usepackage{soul}
\usepackage{subcaption}
\usepackage{hyperref}
\usepackage{enumitem}
\usepackage{mwe,tikz}\usepackage[percent]{overpic}

\usepackage{tabularx}
\newcolumntype{Y}{>{\centering\arraybackslash}X}
\usepackage{xcolor}
\newcommand{\gb}{\textcolor{black}}

\title{Shallow Optical Flow Three-stream CNN\\ for Macro- and Micro-Expression Spotting from Long Videos}
\name{Author(s) Name(s)
}
\address{Author Affiliation(s)}
\name{Gen-Bing Liong$^{\dag}$$^{\star}$ \qquad John See$^{\ddag}$$^{\star}$ \qquad Lai-Kuan Wong$^{\dag}$}
\address{$^{\dag}$ Faculty of Computing and Informatics, Multimedia University, Malaysia\\$^{\ddag}$ 
School of Mathematical and Computer Sciences, Heriot-Watt University Malaysia
\\(${}^{\star}$ Corresponding authors)\vspace{-0.3em}
}

\begin{document}
%
\maketitle
\global\csname @topnum\endcsname 0
\global\csname @botnum\endcsname 0

\begin{abstract}
Facial expressions vary from the visible to the subtle. 
In recent years, the analysis of micro-expressions--- a natural occurrence resulting from the suppression of one's true emotions, has drawn the attention of researchers with a broad range of potential applications. However, spotting micro-expressions in long videos becomes increasingly challenging when intertwined with normal or macro-expressions. In this paper, we propose a shallow optical flow three-stream CNN (SOFTNet) model to predict a score that captures the likelihood of a frame being in an expression interval. By fashioning the spotting task as a regression problem, we introduce pseudo-labeling to facilitate the learning process. We demonstrate the efficacy and efficiency of the proposed approach on the recent MEGC 2020 benchmark, where state-of-the-art performance is achieved on CAS(ME){$^2$} with equally promising results on SAMM Long Videos. 
\end{abstract}
\begin{keywords}
Micro-expression, macro-expression, spotting, optical flow, shallow CNN
\end{keywords}
\vspace{-0.5em}
\section{Introduction}
\vspace{-0.5em}
In most naturalistic scenarios, spontaneous facial expressions could occur at varying degrees of intensities and brevity -- from the visible to the subtle. These occurrences of \emph{macro-expressions} and \emph{micro-expressions} could coexist or occur in isolation. Micro-expressions, which typically lasts between 1/25 to 1/5 second at rather low intensities, occur when a person attempts to conceal his or her genuine emotions in a high-stake situation~\cite{ekman2009telling}. On the other hand, macro-expressions are easier to identify even without proper training as the duration is longer at higher intensities. Recent advances in deep learning have witnessed a widespread popularity in the recognition task while efforts in the spotting task, especially on long ``untrimmed'' videos, remain subdued~\cite{oh2018survey}. As such, the micro-expression community has recently organized the 3rd MEGC Workshop (MEGC2020) \cite{li2020megc2020} to challenge researchers towards spotting macro- and micro-expression in long videos. Generally, facial expressions undergo three distinct phases: \emph{onset}, \emph{apex}, and \emph{offset}. As accurately described in \cite{valstar2011fully}, onset occurs when facial muscles begin contracting; apex is the phase where the facial action is at its peak intensity; offset signifies the muscles going back to neutral state. This paper highlights the task of spotting both macro- and micro-expression sequences, \emph{i.e.} from onset to offset.
\begin{figure}[t!]
    \centering
    \includegraphics[scale=0.45]{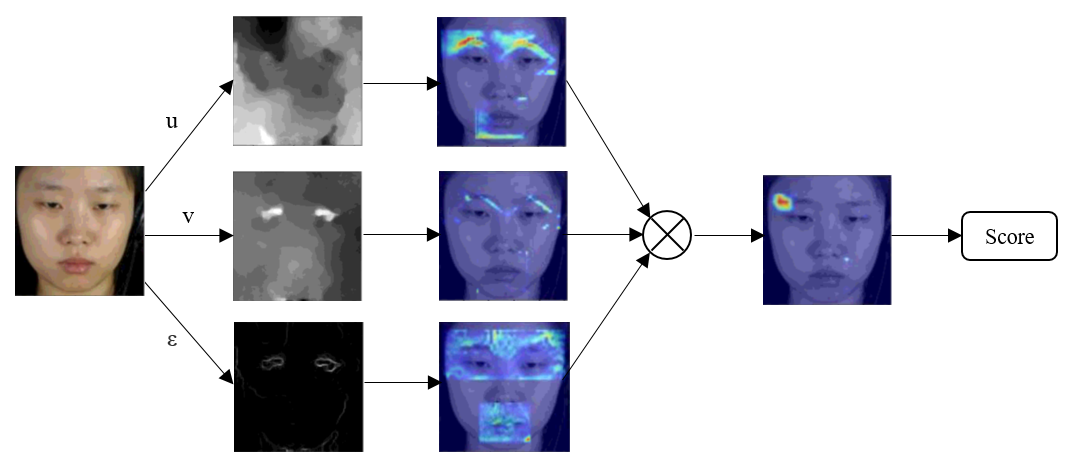}
    \vspace{-1.5em}
    \caption{Intuitively, the optical flow components used in each of the three streams capture different salient motion information for uncovering macro- and micro-expressions.}
    \vspace{-1.5em}
    \label{fig:overview}
\end{figure}

Early works by \cite{shreve2014automatic} and \cite{moilanen2014spotting} notably laid the fundamental mechanism of the task; the latter in particular employed LBP as the feature descriptor with {$\chi$}{$^2$}-distance used for feature difference (FD) analysis between two frames in a fixed duration. The micro-expression is determined if the frame's feature vector is above the threshold set for peak detection. 
Most works utilise established pre-processing techniques involving landmark detection~\cite{cristinacce2006feature, asthana2013robust}, region masking \cite{shreve2014automatic, liong2016automatic}, and emphasis on specific facial regions via ROI selection \cite{liong2015automatic, li2018ltp, verburg2019micro}. 
\begin{figure*}[t!]
    \centering
    \begin{overpic}[scale=1.00]{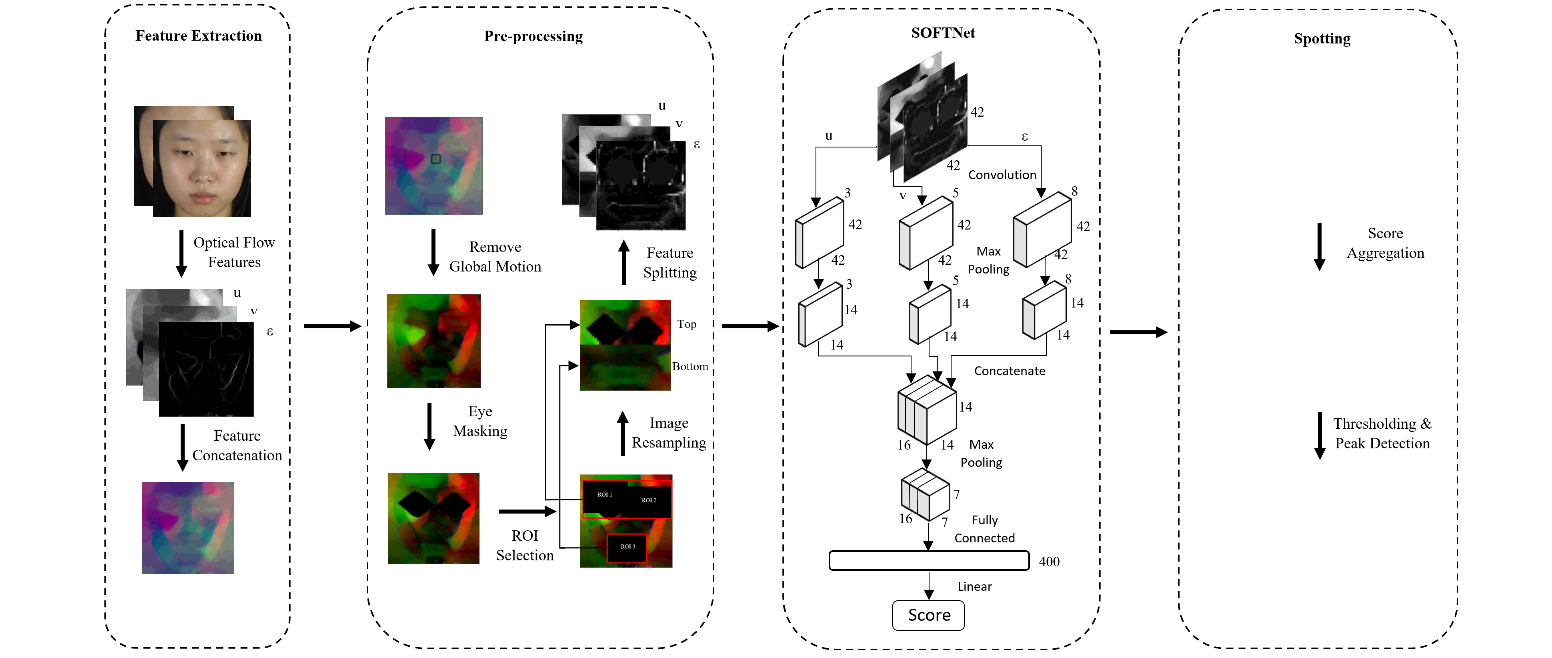}
        \put(75.5,28.8){\includegraphics[width=8.7em,height=4.2em]{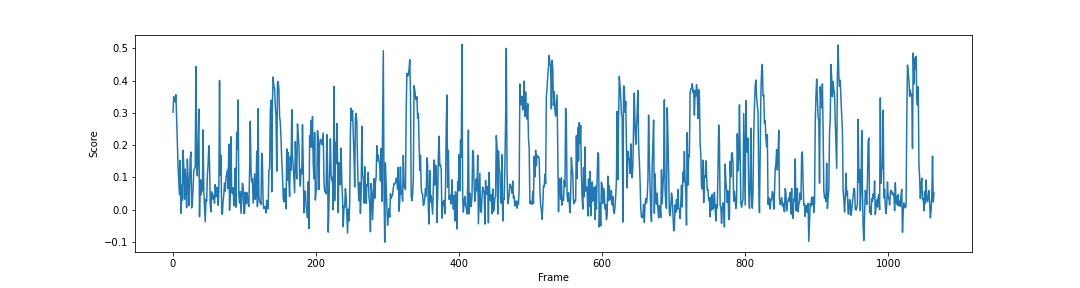}}  
        \put(75.5,16.3){\includegraphics[width=8.7em,height=4.2em]{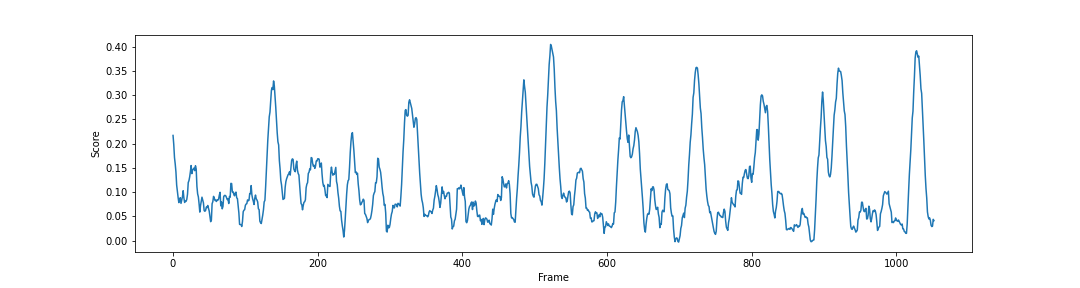}}  
        \put(75.5,4){\includegraphics[width=8.7em,height=4.2em]{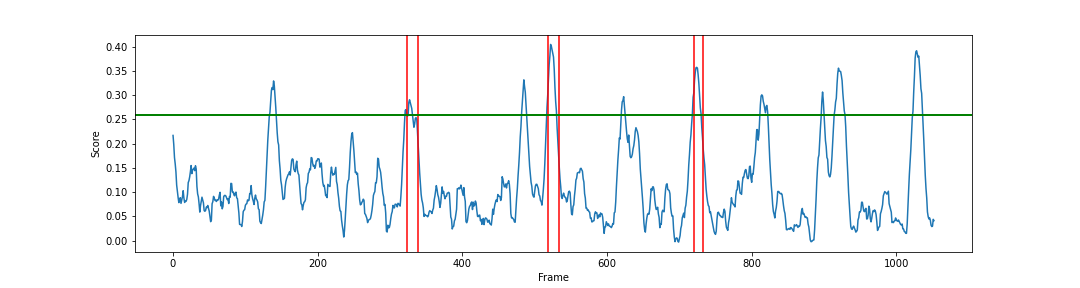}}  
    \end{overpic}
    \vspace{-0.7em}
    \caption{Framework of the proposed approach and its four phases.}
    \label{fig:proposed_framework}
    \vspace*{-1.4em}
\end{figure*}

Motion-based approaches can characterize the subtle movements on the face. Shreve et al. \cite{shreve2014automatic} first introduced optical strain (a derivative of optical flow) to analyze subtle motion changes based on the elastic deformation of facial skin tissue. The amount of strain observed across time (by summing its magnitudes) at different facial regions is considered. The baseline method for MEGC2020, MDMD \cite{wang2017main} encodes the maximal difference magnitude along the main direction of motion to predict if a macro- or micro-expression is present. Meanwhile, \cite{liong2015automatic} constructed optical strain features for apex spotting.
More recently, a few works~\cite{tran2020micro, verburg2019micro} have begun to adopt deep learning methods for spotting. \cite{tran2020micro} experimented with CNN and RNN models under an alternative benchmarking strategy while~\cite{verburg2019micro} fed pre-computed HOOF features into an RNN which is fashioned to spot short intervals with likely micro-movements.  
To address issues such as shortage of micro-expression samples and over-complex models, \cite{liong2019shallow} proposed to use shallow convolutional networks with multiple streams of input information. Their work however, was catered for the recognition task.

Inspired by \cite{liong2019shallow}, we hypothesize that such kind of models can be trained to alleviate the insufficiency of data, but concurrently harnessing the benefits of motion information. To achieve this, we fashion the spotting task as a \emph{regression} problem that predicts how likely a frame belongs to a macro- or micro-expression. In the core, we build a shallow optical flow three-stream CNN (SOFTNet) to capture the relevant features from different optical flow components. 
The contributions of this paper are summarized as follows:
\begin{enumerate}[topsep=2pt,itemsep=-1ex]
  \item We propose a multi-stream shallow network infused with optical flow inputs to regress a score for spotting.
  \item We present a new automatic way of pseudo-labeling frames to enable the training of a regression model.
  \item We demonstrate the efficacy of the proposed approach in terms of F1-score and computational time on the MEGC2020 benchmark, achieving state-of-the-art results on CAS(ME)$^2$.
  \item We re-explore the viability of a detection metric which provides a fairer and more consistent measure for locating both macro- and micro-expression occurrences.
\end{enumerate}

\vspace{-1em}
\section{Proposed Framework}
\vspace{-0.5em}
The proposed framework is illustrated in Figure \ref{fig:proposed_framework}. 
This section discusses the four phases of the framework: initial feature extraction of optical flow components, a series of pre-processing steps, feature learning with the SOFTNet regression network, and finally the expression spotting procedure.

\vspace{-0.70em}
\subsection{Feature Extraction}
\vspace{-0.40em}
The prevalent use of optical flow features in several works in micro-expression analysis \cite{shreve2014automatic, liong2019shallow, khor2019dual} have shown the usefulness of spatio-temporal motion information. To normalize the face resolution, the facial region in each frame is cropped and resized to $128\times128$ pixels. 
Cropping was performed using the Dlib toolbox~\cite{king2009dlib} after the 68 landmark points were detected from the first (reference) frame of each raw video. 

Subsequently, \emph{optical flow} features are computed from two frames, \emph{i.e.} current frame $F_i$ and the $k$-th frame from $i$, $F_{i+k}$, where $k$ is half of the average length of an expression. Horizontal and vertical components, $\mathbf{u}$ and $\mathbf{v}$ respectively are computed using TV-L1 optical flow method \cite{perez2013tv}. Additionally, \emph{optical strain}, which is adopted from infinitesimal strain theory, captures the subtle facial deformation from optical flow components~\cite{shreve2014automatic}. It can be defined as follows:
\vspace{-0.4em}
\begin{equation}
	\epsilon =  \begin{bmatrix}
	\epsilon_{xx} = \frac{\delta u}{\delta x} & \epsilon_{xy} = \frac{1}{2}(\frac{\delta u}{\delta y} + \frac{\delta v}{\delta x}) \\
    \epsilon_{yx} = \frac{1}{2}(\frac{\delta v}{\delta x} + \frac{\delta u}{\delta y}) & \epsilon_{yy} = \frac{\delta v}{\delta y}
	\end{bmatrix} 
    \label{bimatrix}
    \vspace{-0.4em}
\end{equation}
where $\epsilon_{xx}$ and $\epsilon_{yy}$ indicate normal strain components while $\epsilon_{xy}$ and $\epsilon_{yx}$ indicate shear strain components. The optical strain magnitude $\epsilon$, can be computed as:
\setlength{\belowdisplayskip}{2pt} \setlength{\belowdisplayshortskip}{0pt}
\setlength{\abovedisplayskip}{2pt} \setlength{\abovedisplayshortskip}{0pt}
\begin{equation}
	|\epsilon| = \sqrt[]{\epsilon_{xx}^2 + \epsilon_{yy}^2 + \epsilon_{xy}^2 + \epsilon_{yx}^2}.
	\label{eq:strain_mag}
\end{equation}

\noindent These three components ($\mathbf{u}$, $\mathbf{v}$, and $\epsilon$) 
represent the input data for the model learning phase. 

\vspace{-0.4em}
\subsection{Pre-processing}
Prior to the learning phase, a series of pre-processing steps are introduced 
to ensure consistency of the data before model learning. Motivated by the work of \cite{zhang2020spatio}, we take the landmark position of the nose region with five pixels margin to eliminate the global head motion for each frame.


Then, we omit the left and right eye regions since optical flow features are highly sensitive to eye blinking \cite{liong2016automatic}. Following this work, a polygon (with extra margin of 15 pixels along the \gb{height and width}) is applied to mask out these eye regions. Next, the area bounded by three regions \gb{(with extra margin of 12 pixels)}: (1) left eye and left eyebrow; (2) right eye and right eyebrow; (3) mouth, are selected on the basis that they typically contain significant movements \cite{liong2015automatic, liong2016automatic}. \gb{The regions obtained are re-sampled to form an image of size $42\times42$ (height, width) which retains the important features. Specifically, the top part is obtained by horizontally stacking the resized ROI 1 and ROI 2 of similar size $21\times21$, while the bottom part comes from the resized ROI 3 of size $21\times42$.}

\vspace{-0.85em}
\subsection{Shallow Optical Flow Three-stream CNN}
\vspace{-0.3em}
\begin{figure}[t!]
    \hspace{-2.0em}
    \centering
    \includegraphics[scale=0.5]{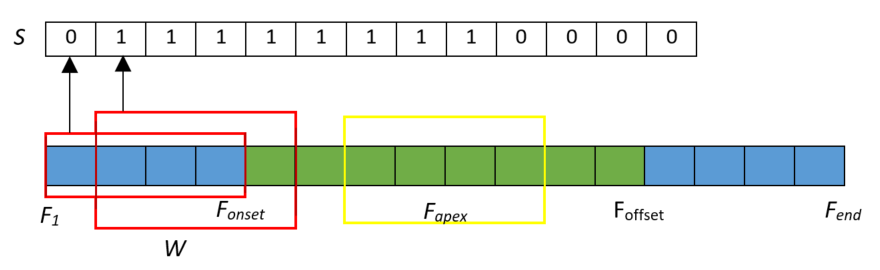}
    \vspace{-0.75em}
    \caption{Pseudo-labeling in a video using sliding window approach. Blue frames are not part of the expression interval while green frames are within the interval.}
    \label{fig:labeling}
    \hspace{-2.0em}
    \vspace{-2.5em}
\end{figure}

\noindent \textbf{SOFTNet.} Motivated by the architecture in \cite{liong2019shallow}, we propose SOFTNet with these further considerations: (1) The convolutional layer applies a $5\times5$ filter rather than $3\times3$ to increase the receptive field coverage to accommodate macro-expressions; (2) A regression output layer added to predict the score of each frame corresponding to its likelihood of being involved in the interval of expression. Intuitively, optical flow features are typically not as  significant at the middle frames of a sequence (yellow color window in Fig.~\ref{fig:labeling}) as compared to frames nearer to the $F_{onset}$ and $F_{offset}$ hence it is desirable to regress a high score for peak detection. 

SOFTNet is a three-stream (very) shallow architecture where each stream consisting of a single convolutional layer with 3, 5, and 8 filters respectively, followed by a max-pooling layer to reduce the feature map size. The feature maps from each stream are then stacked channel-wise to combine the features, with another max-pooling layer thereafter. Finally, it flattens out to a 400-node layer, fully connected to a single output score via linear activation. Specifically, the learned model $\mathcal{M}$ takes in the three optical flow components $\mathbf{u}$, $\mathbf{v}$, and $\epsilon$ of the $i$-th frame as input to each stream and predicts a spotting confidence score $\hat{s}_i$. Separate models are learned to spot macro-expressions ($\alpha$) and micro-expressions ($\beta$), \emph{i.e.} $\hat{s}_{i,\phi} =\mathcal{M}_\phi(\mathbf{u}_i,\mathbf{v}_i,\epsilon_i)$ where $\phi=\{\alpha,\beta\}$.



\noindent \textbf{Pseudo-labeling.} Since ground-truth labels only provide the onset, offset and apex frame indices, to realize a sliding window mechanism, we need to create labels for each window position. To label the frame in videos, the sliding window, $W_j$ at the $j$-th position with length $k$\footnote{$k=(N+1)/2$ is half the average length of expression in each dataset.
} corresponding to interval $[F_i, F_i+k-1]$, is scanned across each video. We impose a pseudo-labeling function $g$ (for Heaviside step function, $g(IoU) = 0$ if $IoU \leq 0$, else $g(IoU) = 1$), which to determine the score $s$ for each $j$-th window calculated from the $IoU$ between $W$ and $\mathcal{E}$:
\setlength{\belowdisplayskip}{4pt} 
\setlength{\abovedisplayskip}{7pt} 
\begin{equation}
\begin{gathered}
    IoU = \frac{W \bigcap \mathcal{E}}{W \bigcup \mathcal{E}} \\
    \text{where} \hspace{0.4em} \mathcal{E} = [F_{onset}, F_{offset}]
\end{gathered}
\end{equation}
Finally, the pseudo-label set $S = \{s_{i,\phi}$ for $i=1,\ldots,F_{end}-k\}$ which represents the labels (of $\phi$) for the SOFTNet inputs, is obtained, as illustrated in Figure \ref{fig:labeling} example. Other pseudo-labeling functions such as linear function and step function were found to be less desirable after experiments. 


\noindent \textbf{Training configurations.} In our experiments, we applied SGD with learning rate 5 x 10$^{-4}$ with the number of epochs set to 10. Since the dataset is highly imbalanced, we opt to sample 1 of every 2 non-expression frames, similar to the strategy in~\cite{li2018ltp}. Data augmentation including horizontal flip, Gaussian blur ($7\times 7$), and adding random Gaussian noise ($\mathcal{N}(0,1)$), is performed during micro-expression training only to address the small sample size problem. 

\vspace{-1.0em}
\subsection{Spotting}
\vspace{-0.5em}
The predicted score of each frame is aggregated as:
\setlength{\belowdisplayskip}{4pt} \setlength{\belowdisplayshortskip}{0pt}
\setlength{\abovedisplayskip}{4pt} \setlength{\abovedisplayshortskip}{0pt}
\begin{equation}
\begin{gathered}
	\hat{s}_{i,\phi} = \frac{1}{2k+1}\sum^{i+k}_{j=i-k}\hat{s}_{j,\phi}\hspace{0.4em}
	\displaystyle \text{for} \hspace{0.4em} i \! = F_{1} \! + \! k,\! \ldots \!, \!F_{end} \!- \! k
\end{gathered}
\end{equation}
whereby the predicted scores from $k$ frames before until $k$ frames after the current $i$-th frame are averaged for smoothing purpose. Intuitively, each frame now represents a potential interval of expression by accumulation of confidence scores.

Finally, we employ the standard threshold and peak detection technique of \cite{moilanen2014spotting} to spot the peaks in each video where the threshold is defined as:
\begin{equation}
    \label{eq:peak_threshold}
    T = \hat{S}_{mean} + p \times (\hat{S}_{max} - \hat{S}_{mean})
\end{equation}
where $\hat{S}_{mean}$ and $\hat{S}_{max}$ are the average and maximum predicted score over the entire video, and $p$ is a tuning parameter in the range of $[0, 1]$. As shown in Figure \ref{fig:proposed_framework} spotting phase, the green line (bottom row) is the threshold and red lines indicate a few intervals of expressions. A peak frame $s_{P,\phi}$ is spotted by finding a local maxima (with minimum distance of $k$ between peaks) and extending by $k$ frames to obtain the spotted interval $\hat{\mathcal{E}}_{\phi}=[s_{P,\phi}-k, s_{P,\phi}+k]$ for evaluation.
\vspace{-0.2em}
\begin{table*}[t]
    \centering
    \caption{Comparison between the proposed approach against baseline and state-of-the-art methods in F1-score}
    \vspace{-0.75em}
        \begin{tabularx}{\linewidth}{| l | Y | Y | Y | Y | Y | Y |}
        \hline
        \multicolumn{1}{|c|}{Dataset} & \multicolumn{3}{c|}{CAS(ME){$^2$}} & 
        \multicolumn{3}{c|}{SAMM Long Videos} 
        \\
        \hline
        \multicolumn{1}{|c|}{Methods} &
        Macro & Micro& Overall & Macro & Micro & Overall \\
        \hline
        Baseline \cite{he2019spotting} & 0.1196 & 0.0082 & 0.0376 & 0.0629 & 0.0364 & 0.0445 \\ 
        Gan et al \cite{li2020megc2020} & 0.1436 & 0.0098 & 0.0448 & - & - & - \\    
        Pan \cite{li2020megc2020} & - & - & 0.0595 & - & - & 0.0813 \\
        Zhang et al. \cite{zhang2020spatio} & 0.2131 & 0.0547 & 0.1403 & 0.0725 & 0.1331 & 0.0999 \\
        Yap et al. \cite{yap2019samm}  & - & - & - & \textbf{0.4081} & 0.0508 & \textbf{0.3299} \\ \hline
        Ours (W/o SOFTNet)  & 0.1615 & \textbf{0.1379} & 0.1551 & 0.1463 & 0.1063 & 0.1293 \\
        \textbf{Ours (With SOFTNet)}  & \textbf{0.2410} & 0.1173 & \textbf{0.2022} & 0.2169 & \textbf{0.1520} & 0.1881 \\
        \hline 
        \end{tabularx}
        \vspace{-0.25em}
        \label{table:benchmarking}
\end{table*}

\vspace{-0.8em}
\section{Experiments}
\vspace{-0.7em}
To demonstrate the effectiveness of the proposed framework, we conduct extensive experiments on the MEGC 2020 spotting benchmark. It is important to note that the SOFTNet models are implemented separately (\emph{i.e.} training and inference) for both macro- and micro-expressions. 
To encourage community usage, the code is publicly available \footnote{Link: \url{https://github.com/genbing99/SoftNet-SpotME}}.

\vspace{-0.6em}
\subsection{Evaluation Details}
\vspace{-0.4em}
\noindent \textbf{Datasets.} Two benchmark datasets, namely CAS(ME){$^2$} \cite{qu2016cas} and SAMM Long Videos \cite{yap2019samm} are used. Briefly, CAS(ME){$^2$} contains 98 long videos consisting of 300 macro-expressions and 57 micro-expressions captured from 22 subjects; SAMM Long Videos is an extension of SAMM \cite{davison2016samm}, one of the most culturally diverse datasets in this domain, with 147 long videos (343 macro-movements, 159 micro-movements) elicited from 32 subjects. However, a small number (10) of macro-expression samples were discarded due to the ambiguous onset annotation. In spite of that, both datasets were fully annotated with onset, apex, and offset by professional coders.

\noindent \textbf{Performance Metric.} We benchmark our proposed approach against recent works from MEGC 2020~\cite{li2020megc2020}, adopting the similar F1-score metric for both macro- and micro-expression spotting. Besides, we propose the use of Average Precision over different Intersection over Union (IoU) thresholds from 0.5 to 0.95 with a step size of 0.05 (denoted as \textbf{AP@[.5:.95]}), a popular metric used in MS COCO \cite{girshick2015fast}, to provide a more consistent measure of the quality of the spotting result. 

\noindent \textbf{Settings.} Leave-one-subject-out (LOSO) cross-validation is applied to ensure all samples are evaluated. For peak detection, 
we empirically select $p=0.55$ for SOFTNet and $p=0.5$ for without SOFTNet. Parameter $k$ is computed to be $\{6, 18\}$ for CAS(ME){$^2$} and $\{37, 174\}$ for SAMM (smaller value for micro, larger value for macro).

\vspace{-0.6em}
\subsection{Results and Discussions}
\vspace{-0.3em}
\begin{table}[t!]
\vspace{-1.3em}
\footnotesize
    \caption{Detailed result of the proposed SOFTNet approach}
    \vspace{-0.75em}
    \begin{tabularx}{\linewidth}{| l | Y | Y | Y | Y | Y | Y |}
        \hline
        \multicolumn{1}{|c|}{Dataset} & \multicolumn{3}{c|}{CAS(ME){$^2$}} & 
        \multicolumn{3}{c|}{SAMM Long Videos} 
        \\
        \hline
        Expression &
        Macro & Micro & Overall & Macro & Micro & Overall \\
        \hline
        Total & 300 & 57 & 357 & 333 & 159 & 492 \\ 
        TP & 90 & 20 & 110 & 68 & 38 & 106 \\ 
        FP & 357 & 264 & 621 & 226 & 303 & 529 \\    
        FN & 210 & 37 & 247 & 265 & 121 & 386 \\
        Precision & 0.2013 & 0.0704 & 0.1505 & 0.2313 & 0.1114 & 0.1669 \\
        Recall & 0.3000 & 0.3509 & 0.3081 & 0.2042 & 0.2390 & 0.2154 \\
        F1-Score & 0.2410 & 0.1173 & 0.2022 & 0.2169 & 0.1520 & 0.1881 \\
        AP@[.5:.95] & 0.0168 & 0.0112 & 0.0140 & 0.0117 & 0.0103 & 0.0110 \\
        \hline 
        \end{tabularx}
        \vspace{-1.5em}
        \label{table:result_details}
\end{table}



Table \ref{table:benchmarking} compares the performance of our proposed approach with the \gb{results (from original publications) of} accepted submissions in MEGC 2020~\cite{li2020megc2020} on both datasets. Our best approach is capable of outperforming other methods on CAS(ME){$^2$} while on the SAMM Long Videos, we achieve the highest F1-score for micro-expressions and is second best for macro-expressions, behind the original dataset authors \cite{yap2019samm}. The control experiment (without SOFTNet and image resampling in pre-processing) determines the spotting score by the sum of the feature map for each frame as suggested in \cite{shreve2014automatic}.
By examining in detail our SOFTNet approach in Table \ref{table:result_details}, 
the amount of TP that we obtained is comparable with other approaches whilst with a much lower FP. The FN is less problematic, but this is proven to be an obstacle in the SAMM Long Videos. The AP@[.5:.95] metric offers a way of equalizing the impact of the half-window length $k$ by considering different IoU levels for matching the intervals. 

\noindent \textbf{Ablation Studies \& Insights.} Table \ref{table:model_compare} compares the SOFTNet against a few popular architectures, showing its superiority across various aspects from accuracy to efficiency. 
Another ablation study on the choice of pseudo-labeling function $g$ shows the unit step (0.2410) performing better than linear (0.2269) and step (0.2092) functions in F1-score. To offer insights into the predictions, we used timeline plots and GradCAM~\cite{selvaraju2017grad} heatmaps to visualize the spotted temporal and spatial locations. Fig. \ref{fig:overview} shows an example of how each stream contributes towards the final outcome. 

\begin{table}[t!]
\small
\vspace{-1.45em}
\centering
\caption{Performance comparison between various network backbones (for CAS(ME){$^2$} macro-expressions)}
\vspace{-0.75em}
        \begin{tabularx}{\linewidth}{| l | Y | Y | Y | Y |}
        \hline
        \multirow{2}{*}{Network} & \multirow{2}{*}{F1-score} & \multicolumn{1}{p{1.1cm}|}{\centering AP@ \newline [.5:.95]} & \multicolumn{1}{p{1.2cm}|}{\centering Inference \newline Time (s)} & \multicolumn{1}{p{1.2cm}|}{\centering Parameter \newline (Million)} 
        \\
        \hline
        \textbf{SOFTNet} & \textbf{0.2410} & \textbf{0.0168} & \textbf{2.7826} & \textbf{0.3148} \\ 
        MobileNetV2 & 0.2152 & 0.0160 & 10.1651 & 2.2631 \\ 
        ResNet-18 & 0.2147 & 0.0150 & 9.3175 & 11.2877 \\ 
        ResNet-50 & 0.1155 & 0.0039 & 22.2178 & 23.5960 \\ 
        VGG-16 & 0.1724 & 0.0095 & 8.2692 & 14.7152 \\ 
        \hline 
        \end{tabularx}
        \vspace{-1.65em}
        \label{table:model_compare}
\end{table}

\vspace{-1.4em}
\section{Conclusion}
\vspace{-0.8em}
This paper proposes a new regression-based strategy towards macro- and micro-expression spotting in long videos by means of a three-stream shallow network based on optical flow information. On the MEGC 2020 benchmark, our approach achieved promising results on both CAS(ME){$^2$} and SAMM Long Videos. No less importantly, we re-introduce the AP@[.5:.95] metric (from object detection) which measures more consistently across both expression types. 
We surmise through findings in this paper that spotting both micro- and macro-expressions demands for innovative modeling of the localized facial transitions and robust peak detection. 

\noindent \textbf{Acknowledgement}: This work is supported in part by Malaysia Ministry of Education FRGS Research Grant\\(Project No: FRGS/1/2018/ICT02/MMU/02/2).
\graphicspath{{./Images/}{./CAM/}}

\clearpage
\section*{Supplemental Notes}

This Supplemental Material provides the justification of our parameter choices, other tested pseudo-labeling functions, and also examples of visualizations in the form of timeline plots and class activation maps of selected samples from the CAS(ME){$^2$} dataset.

\subsection*{A. Parameter Selection}

\subsubsection*{I. Spotting tuning parameter \texorpdfstring{$p$}{p}}
\label{suppmat-pvalue}
For the selection of threshold parameter $p$ used in the peak detection (Eqn. 5 in paper), we tested our proposed approaches empirically by varying $p$ from 0.05 to 0.95 with a step size of 0.05. The result for SOFTNet approach is reported in Table \ref{table:p_threshold} while the approach without SOFTNet is shown in Table \ref{table:p_threshold_without}. We observed that the F1-score is the highest when $p$ is around the middle of the [0, 1] range. 

\begin{table}[b!]
\fontsize{8pt}{10pt}
\selectfont
\centering
\caption{Result for parameter $p$ varying from 0.05 to 0.95 in terms of F1-score of proposed SOFTNet approach}
\vspace{-0.5em}
        \begin{tabularx}{\linewidth}{| Y | Y |  Y | Y | Y | Y | Y |}
        \hline
        \multicolumn{1}{|c|}{Dataset} & \multicolumn{3}{c|}{CAS(ME){$^2$}} & 
        \multicolumn{3}{c|}{SAMM Long Videos} 
        \\
        \hline
        $p$ & Macro & Micro & Overall & Macro & Micro & Overall \\
        \hline
        0.05 & 0.1331 & 0.0239 & 0.0701 & 0.1996 & 0.0632 & 0.1046 \\ 
        0.10 & 0.1609 & 0.0292 & 0.0869 & 0.2061 & 0.0663 & 0.1122 \\ 
        0.15 & 0.1847 & 0.0324 & 0.1029 & 0.2087 & 0.0755 & 0.1239 \\ 
        0.20 & 0.2041 & 0.0384 & 0.1204 & 0.2138 & 0.0845 & 0.1351 \\ 
        0.25 & 0.2179 & 0.0477 & 0.1378 & 0.2138 & 0.0944 & 0.1445 \\ 
        0.30 & 0.2276 & 0.0603 & 0.1558 & 0.2166 & 0.1071 & 0.1558 \\ 
        0.35 & 0.2361 & 0.0697 & 0.1693 & 0.2130 & 0.1158 & 0.1615 \\ 
        0.40 & 0.2403 & 0.0747 & 0.1787 & 0.2148 & 0.1211 & 0.1671 \\ 
        0.45 & \textbf{0.2511} & 0.1933 & 0.0853 & 0.2127 & 0.1279 & 0.1715 \\ 
        0.50 & 0.2484 & 0.1028 & 0.2010 & 0.2165 & 0.1378 & 0.1801 \\ 
        \textbf{0.55} & 0.2410 & 0.1173 & \textbf{0.2022} & \textbf{0.2169} & 0.1520 & \textbf{0.1881} \\ 
        0.60 & 0.2232 & 0.1288 & 0.1947 & 0.2040 & 0.1521 & 0.1818 \\ 
        0.65 & 0.2135 & 0.1401 & 0.1924 & 0.1903 & 0.1606 & 0.1780 \\ 
        0.70 & 0.1966 & 0.1593 & 0.1861 & 0.1900 & \textbf{0.1609} & 0.1783 \\ 
        0.75 & 0.1808 & \textbf{0.1616} & 0.1757 & 0.1811 & 0.1503 & 0.1689 \\ 
        0.80 & 0.1663 & 0.1600 & 0.1647 & 0.1706 & 0.1509 & 0.1630 \\ 
        0.85 & 0.1379 & 0.1569 & 0.1426 & 0.1548 & 0.1212 & 0.1421 \\ 
        0.90 & 0.1149 & 0.1129 & 0.1145 & 0.1486 & 0.1119 & 0.1353 \\ 
        0.95 & 0.0880 & 0.1121 & 0.0930 & 0.1481 & 0.1181 & 0.1374 \\ 
        \hline 
        \end{tabularx}
        \vspace{-0.75em}
        \label{table:p_threshold}
\end{table}

\begin{table}[t!]
\fontsize{8pt}{10pt}
\selectfont
\centering
\caption{Result for parameter $p$ varying from 0.05 to 0.95 in terms of F1-score of proposed approach without SOFTNet}
    \vspace{-0.5em}
        \begin{tabularx}{\linewidth}{| Y | Y | Y | Y | Y | Y | Y |}
        \hline
        \multicolumn{1}{|c|}{Dataset} & \multicolumn{3}{c|}{CAS(ME){$^2$}} & 
        \multicolumn{3}{c|}{SAMM Long Videos} 
        \\
        \hline
        $p$ & Macro & Micro & Overall & Macro & Micro & Overall \\
        \hline
        0.05 & 0.1244 & 0.0537 & 0.0959 & \textbf{0.1771} & 0.0729 & 0.1118 \\ 
        0.10 & 0.1407 & 0.0622 & 0.1110 & 0.1674 & 0.0824 & 0.1173 \\ 
        0.15 & 0.1550 & 0.0754 & 0.1272 & 0.1693 & 0.0963 & 0.1280 \\ 
        0.20 & 0.1641 & 0.0846 & 0.1375 & 0.1692 & 0.1046 & \textbf{0.1339} \\ 
        0.25 & 0.1607 & 0.0789 & 0.1342 & 0.1669 & 0.0986 & 0.1310 \\ 
        0.30 & 0.1591 & 0.0836 & 0.1355 & 0.1592 & 0.1049 & 0.1319 \\ 
        0.35 & \textbf{0.1660} & 0.1020 & 0.1473 & 0.1567 & \textbf{0.1086} & 0.1338 \\ 
        0.40 & 0.1646 & 0.1154 & 0.1504 & 0.1571 & 0.1033 & 0.1325 \\ 
        0.45 & 0.1603 & 0.1293 & 0.1516 & 0.1514 & 0.1028 & 0.1302 \\ 
        \textbf{0.50} & 0.1615 & 0.1379 & \textbf{0.1551} & 0.1463 & 0.1063 & 0.1292 \\ 
        0.55 & 0.1589 & 0.1436 & 0.1550 & 0.1464 & 0.0894 & 0.1232 \\ 
        0.60 & 0.1443 & 0.1472 & 0.1451 & 0.1423 & 0.0826 & 0.1183 \\ 
        0.65 & 0.1447 & \textbf{0.1528} & 0.1466 & 0.1440 & 0.0748 & 0.1170 \\ 
        0.70 & 0.1413 & 0.1250 & 0.1377 & 0.1385 & 0.0736 & 0.1140 \\ 
        0.75 & 0.1281 & 0.1416 & 0.1309 & 0.1399 & 0.0697 & 0.1138 \\ 
        0.80 & 0.1199 & 0.1468 & 0.1254 & 0.1384 & 0.0741 & 0.1151 \\ 
        0.85 & 0.1179 & 0.1509 & 0.1247 & 0.1368 & 0.0766 & 0.1153 \\ 
        0.90 & 0.1100 & 0.1212 & 0.1122 & 0.1342 & 0.0703 & 0.1114 \\ 
        0.95 & 0.1047 & 0.0879 & 0.1014 & 0.1325 & 0.0729 & 0.1115 \\ 
        \hline 
        \end{tabularx}
        \vspace{-0.75em}
        \label{table:p_threshold_without}
\end{table}

\subsubsection*{II. Sliding window length \texorpdfstring{$k$}{k}}
The sliding window length $k$ used in the pseudo-labeling step (Section 2.3 in paper) is determined by obtaining half the average length $N$ of a particular type of expression (micro or macro) for each dataset, \emph{i.e.} $k=(N+1)/2$. Table \ref{table:kvalues} specifies the $k$ values determined from each subset of the data with their actual duration (seconds) given in parenthesis. It is important to note that the $k$ for micro-expression is within the commonly accepted range of 1/25 to 1/5 of a second~\cite{ekman2009telling} while the more noticeable macro-expressions are typically anywhere between 1/2 of a second up to 4 seconds~\cite{Ekman2003EmotionsRR}. Our $k$ values concur with these acceptable duration ranges.
\begin{table}[t!]
\centering
\caption{Sliding window length $k$ used for different datasets and expression types. Actual duration in parenthesis.}
\vspace{-0.5em}
        \begin{tabular}{| c | c | c |}
        \hline
        Dataset & Micro & Macro \\
        \hline
        CAS(ME)$^2$ & \textbf{6} (0.2$s$) & \textbf{18} (0.6$s$)  \\ 
        SAMM Long & \textbf{37} (0.185$s$) & \textbf{174} (0.87$s$) \\ 
        \hline 
        \end{tabular}
        \vspace{-0.5em}
        \label{table:kvalues}
\end{table}

\subsection*{B. Pseudo-labeling}

\begin{figure}[t!]
    \centering 
    \begin{subfigure}[b]{0.2\textwidth}
        \centering
       \includegraphics[width=\textwidth]{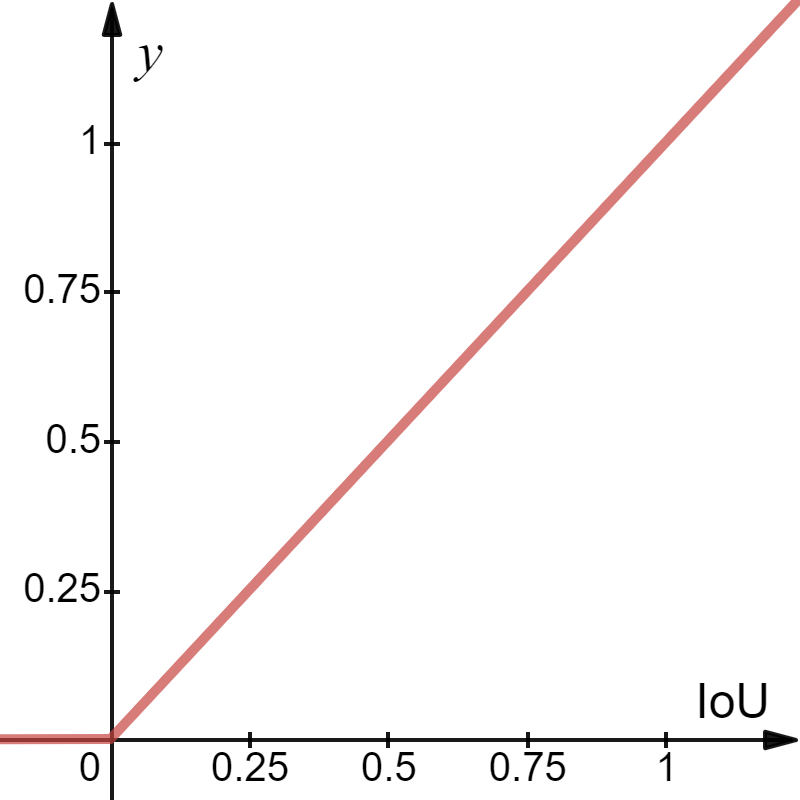}
       \caption{Linear}
       \vspace{1.4em}
       \label{fig:linear_ramp}
    \end{subfigure} \hspace{0.5em}
    \begin{subfigure}[b]{0.2\textwidth}
        \centering
        \includegraphics[width=\textwidth]{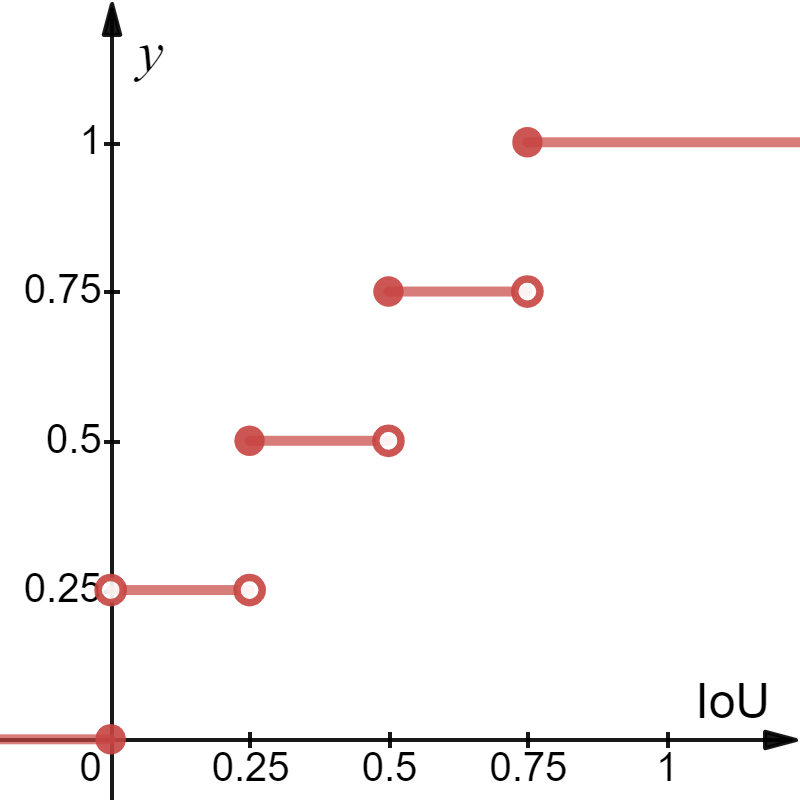}
        \caption{Step Function (4 intervals)}
        \vspace{0.3em}
        \label{fig:step}
     \end{subfigure}
     \begin{subfigure}[b]{0.2\textwidth}
        \centering
        \includegraphics[width=\textwidth]{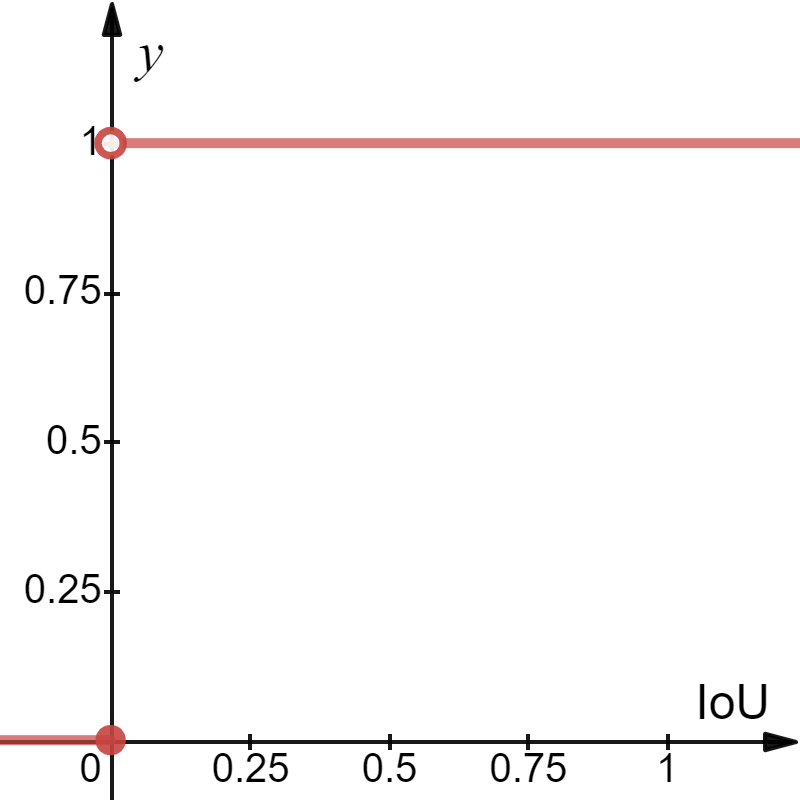}
        \caption{Unit Step}
        \vspace{0.5em}
        \label{fig:unit_step}
     \end{subfigure}
     \vspace{-0.8em}
    \caption{Graph illustration of method for frame labeling}
    \label{fig:labeling_funcs}
    \vspace{-1.2em}
\end{figure}

To our best knowledge, this is the first work that employs a pseudo-labeling function to provide confidence scores for each macro- and micro-expression frame in the video. 
To implement the pseudo-labeling step before training the model, we experimented with a number of common functions to handle the labeling at the boundaries (\emph{i.e.} from the normal frames to expression frames and vice versa) based on IoU defined in Eqn. 3 in the paper. These functions are illustrated in Figure \ref{fig:labeling_funcs}, 
whereby the x-axis is the IoU score between the sliding window, $W$ and the interval of macro- or micro-expressions, $\mathcal{E}$. The equations of the tested functions are as follows:

\vspace{-1em}
\begin{flalign}
  \text{Linear:} && g(IoU) = IoU &&
\end{flalign} 
\vspace{-1em}
\begin{flalign} \text{Step Function:} &&
  g(IoU) =
  \begin{cases}
  0 & \text{if $IoU \leq 0$} \\
  0.25 & \text{if $0 < IoU < 0.25$} \\
  0.5 & \text{if $0.25 \leq IoU < 0.5$} \\
  0.75 & \text{if $0.5 \leq IoU < 0.75$} \\
  1 & \text{if $IoU \geq 0.75$} \\
  \end{cases} &&
\end{flalign} 
\vspace{-2em}
\begin{flalign} \text{Unit Step:} &&
  g(IoU) =
  \begin{cases}
  0 & \text{if $IoU \leq 0$} \\
  1 & \text{if $IoU > 0$} \\
  \end{cases} &&
\end{flalign} 
\vspace{-1em}

\begin{table}[ht!]
\centering
\small
\caption{Comparison of frame labeling methods with parameter $p$ = 0.55 on CAS(ME){$^2$} macro-expression}

        \begin{tabularx}{\linewidth}{| c | Y | Y | Y | Y | Y |}
        \hline
        Method & TP & FP & FN & F1-score & \multicolumn{1}{p{1.1cm}|}{\centering AP@ \newline [.5:.95]} 
        \\
        \hline
        Linear & 91 & 411 & 209 & 0.2269 & 0.0172 \\  \hline
        Step Function & 82 & 402 & 218 & 0.2092 & 0.0125 \\  \hline
        \textbf{Unit Step} & 90 & 357 & 210 & \textbf{0.2410} & 0.0168 \\ 
        \hline 
        \end{tabularx}
        \vspace{-0.6em}
        \label{table:frame_label_compare}
\end{table}

We perform an ablation study on these different functions to better ascertain their influence towards spotting when we intend to model the task as a regression problem. In our limited testing (during time of writing), we observe that the unit (Heaviside) step function method outperforms the rest in terms of F1-score when experimented on the CAS(ME){$^2$} macro-expression subset, although the linear function appears to be marginally better in the AP@[.5:.95] metric. 

In future, this opens up further research into the best possible way to characterize the expression boundaries. Past research have explored the use of expression states~\cite{kim2016micro} in micro-expressions and the recognition of temporal phases \cite{valstar2011fully} in expressions.

\vspace{-1em}
\subsection*{C. Visualization}

\vspace{-0.4em}
\subsubsection*{I. Timeline plot for long video spotting}
\vspace{-0.4em}

\begin{figure*}[t]
    \centering
    \subfloat[$\texttt{s23\_disgust2}$]{
    	\label{subfig:long_video_spot_success}
    	\includegraphics[width=1.0\textwidth]{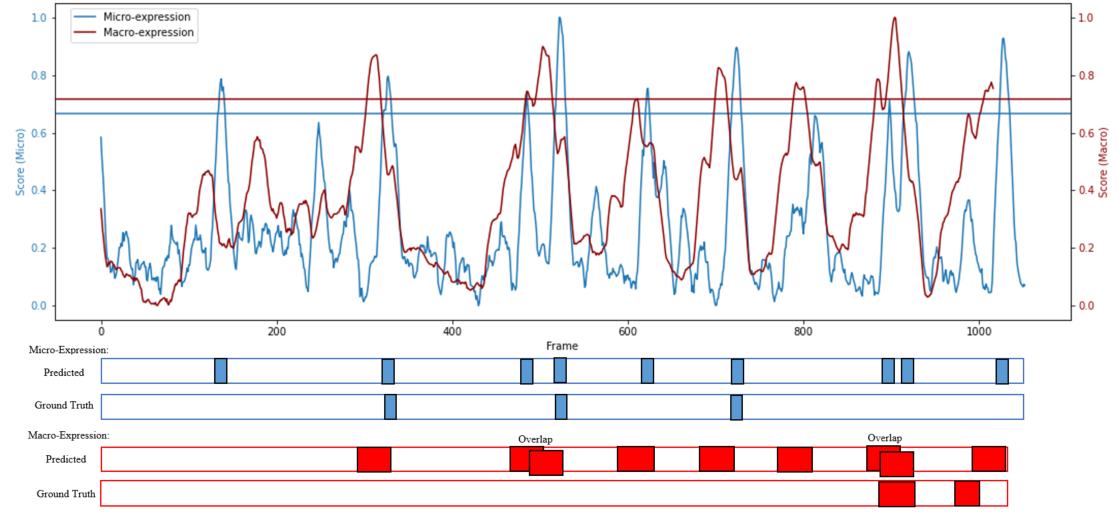} }
    \hfill
    \vspace{1.0em}
    \subfloat[$\texttt{s15\_disgust2}$]{
    	\label{subfig:long_video_spot_fail}
    	\includegraphics[width=1.0\textwidth]{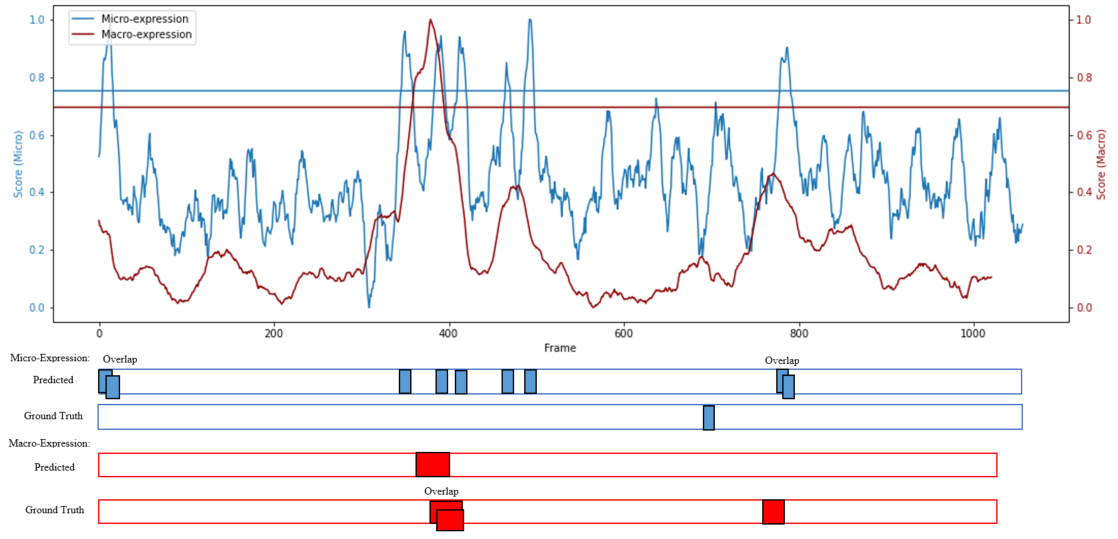} }
    \caption{Visualization of macro- and micro-expression samples in long videos}
    \label{fig:long_video_spot}
\end{figure*}

To meaningfully demonstrate the outcome of the spotting task in a long video, we build timeline plots for a closer examination into success and failure cases. Figure \ref{fig:long_video_spot} depicts the normalized spotting confidence score plots of both macro- and micro-expressions in two sample long videos (both contain a variety of micro- and macro-expressions from CAS(ME){$^2$}). The x-axis denotes the frame number of the video, while the y-axis denotes the normalized confidence scores. The horizontal lines represent the corresponding thresholds ($T$) for peak detection. Below the plot, the predicted interval $\hat{\mathcal{E}}_{\phi}$ and ground truth interval $\mathcal{E}_{gt}$ are provided; some samples are overlapping due to the annotation of \emph{compound} expressions, \emph{i.e.}, a combination of two emotion categories around the same time. 

The examples in Figure \ref{fig:long_video_spot} show that the proposed SOFTNet approach is capable of spotting multiple intervals from long videos. The $\texttt{s23\_disgust2}$ video contains many FPs for macro-expressions while the three micro-expressions were correctly identified. In the $\texttt{s15\_disgust2}$ video, the single micro-expression proved to be very difficult to spot whereas the main macro-expression (the top peak) was obvious and could be spotted by our model. 

\vspace{-0.9em}
\subsubsection*{II. Class activation maps}

\begin{figure}[b!]
    \centering
    \begin{subfigure}[b]{0.2\textwidth}
        \includegraphics[width=0.8\textwidth]{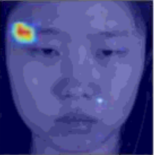}
        \captionsetup{justification=centering}
        \vspace{-0.4em}
        \caption{Disgust\\($\texttt{s15\_disgust1\_1}$)}
        \label{fig:disgust}
    \end{subfigure}
    \hspace{0.3em}
    \begin{subfigure}[b]{0.2\textwidth}
        \includegraphics[width=0.8\textwidth]{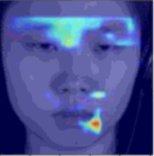}
        \captionsetup{justification=centering}
        \vspace{-0.4em}
        \caption{Anger\\($\texttt{s15\_anger2\_2}$)}
        \label{fig:anger}
    \end{subfigure}
    \hspace{0.3em}
    \begin{subfigure}[b]{0.2\textwidth}
        \includegraphics[width=0.8\textwidth]{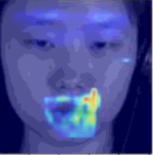}
        \captionsetup{justification=centering}
        \vspace{-0.4em}
        \caption{Happy\\($\texttt{s15\_happy1\_2}$)}
        \label{fig:happy}
    \end{subfigure}
    \vspace{-0.6em}
    \caption{GradCAM visualizations of class activation maps for the proposed SOFTNet on CAS(ME){$^2$} macro-expression.}
    \label{fig:gradcam}
    \vspace{-1em}
\end{figure}

To understand where the SOFTNet model is ``looking'' at in each frame (particularly, at the onset of an expression), we apply Gradient-weighted Class Activation Mapping (GradCAM)~\cite{selvaraju2017grad} to visualize how the model arrived at the predicted decision during inference. The gradients flowing into the final channel-wise concatenated layer produces the coarse localization maps shown in Figure \ref{fig:gradcam}. Note that the resolution of the heatmap is not high due to the $3\times$ upsampling factor (from $14\times 14$ to input size of $42\times 42$). However, it suffices to highlight the important regions in the face area that contribute towards the final spotting score prediction. In these examples, we have picked example frames that have highly confident predicted scores. 

From Figure \ref{fig:gradcam}, the heatmap indicates that the network is able to learn from facial regions that correspond closely to the associated Action Unit (AU) triggered by the specific expression: (Figure \ref{fig:disgust}, \ref{fig:anger}) \textit{Brow Lower} (AU 4) is activated when disgust and anger occurs; (Figure \ref{fig:anger}) \textit{Lip Pressor} (AU 24) is accentuated when anger occurs; (Figure \ref{fig:happy}) \textit{Jaw Sideways} (AU 30) typically indicates a happy emotion.
\vspace{-1em}

\clearpage
\bibliographystyle{IEEEtran}

\bibliography{references}

\end{document}